\documentclass[11pt,a4paper]{article}
\usepackage[utf8]{inputenc}
\usepackage{amssymb}
\usepackage{amsmath}
\usepackage{algorithm}
\usepackage[noend]{algpseudocode}

\makeatletter
\def\BState{\State\hskip-\ALG@thistlm}
\makeatother

\usepackage{bm}
\usepackage{lmodern}
\usepackage{multirow}
\usepackage{float}
\usepackage{subcaption}
\usepackage{mathtools, nccmath}
\usepackage[hyperref]{acl2018}
\usepackage{times}
\usepackage{latexsym}
\usepackage{array}
\usepackage{url}
\usepackage{acl2018}

\usepackage{geometry}

\usepackage{sectsty}
\makeatletter
\newcommand*{\centerfloat}{%
  \parindent \z@
  \leftskip \z@ \@plus 1fil \@minus \textwidth
  \rightskip\leftskip
  \parfillskip \z@skip}
\makeatother
\algdef{SE}[VARIABLES]{Variables}{EndVariables}
   {\algorithmicvariables}
   {\algorithmicend\ \algorithmicvariables}
\algnewcommand{\algorithmicvariables}{\textbf{global variables}}
\DeclareMathOperator*{\argmax}{\arg\!\max}

\newcommand{\eor}{$<$/R$>$ }

\title{Multi-representation Ensembles and Delayed SGD Updates Improve Syntax-based NMT}
  
\author{Danielle Saunders$^\dag$ \and Felix Stahlberg$^\dag$ \and Adri\`a de Gispert$^{\ddagger\dag}$ \and Bill Byrne$^{\ddagger\dag}$ \\
\\
    $^\dag$Department of Engineering, University of Cambridge, UK  \\
\\
    $^\ddagger$SDL Research, Cambridge, UK}

\date{}
\aclfinalcopy

\begin{document}

\maketitle
\begin{abstract}
We explore strategies for incorporating target syntax into Neural Machine Translation. We specifically focus on syntax in ensembles containing multiple sentence representations. We formulate beam search over such ensembles using WFSTs, and describe a delayed SGD update training procedure that is especially effective for long representations like linearized syntax. Our approach gives state-of-the-art performance on a difficult Japanese-English task.\end{abstract}

\section{Introduction}
Ensembles of multiple NMT models consistently and significantly
improve over single models \cite{garmash2016ensemble}. Previous work
has observed that NMT models trained to generate target syntax can
exhibit improved sentence structure \cite{aharoni2017towards,eriguchi2017learning} 
relative to those trained on plain-text, while
plain-text models produce shorter sequences and so may encode lexical
information more easily \cite{nadejde2017predicting}. We hypothesize
that an NMT ensemble would be strengthened if its component models
were complementary in this way.  However, ensembling often requires
component models to make predictions relating to the same output
sequence position at each time step. Models producing different
sentence representations are necessarily synchronized to enable this. We propose an approach to decoding ensembles of models generating different
representations, focusing on models generating syntax.

As part of our investigation we suggest strategies for practical NMT
with very long target sequences. These long sequences may
arise through the use of linearized constituency trees and can be
much longer than their plain byte pair encoded (BPE) equivalent representations (Table
\ref{tab:represent-lens}). Long sequences make training more difficult
\cite{bahdanau15jointly}, which we address with an adjusted training procedure for the
Transformer architecture \cite{vaswani2017attention}, using delayed
SGD updates which accumulate gradients over multiple batches. We also
suggest a syntax representation which results in much shorter
sequences.
\subsection{Related Work}
\citet{nadejde2017predicting} perform NMT with syntax annotation in the form of Combinatory Categorial Grammar (CCG) supertags. \citet{aharoni2017towards} translate from source BPE into target linearized parse trees, but omit POS tags to reduce sequence length. They demonstrate improved target language reordering when producing syntax. \citet{eriguchi2017learning} combine recurrent neural network grammar (RNNG) models \citep{dyer2016recurrent} with attention-based models to produce well-formed dependency trees. \citet{wu2017sequence} similarly produce both words and arc-standard algorithm actions \cite{nivre2004incrementality}. 

Previous approaches to ensembling diverse models focus on model inputs. \citet{hokamp2017ensembling} shows improvements in the quality estimation task using ensembles of NMT models with multiple input representations which share an output representation. \citet{garmash2016ensemble} show translation improvements with multi-source-language NMT ensembles.

\begin{table*}
\centering
\small

\begin{tabular}{>{\raggedright\arraybackslash}p{2.4cm}p{9.5cm}c}
  Representation &Sample  & Mean length\\
  \hline
(1) Plain-text & No complications occurred & 27.5 \\
(2) Linearized tree & (ROOT (S (NP (DT No ) (NNS complications ) ) (VP (VBD occurred ) ) ) )& 120.0 \\
(3) Derivation  & ROOT$\rightarrow$S ; S$\rightarrow$NP VP ; NP$\rightarrow$DT NNS ; DT$\rightarrow$No ; NNS$\rightarrow$complications ; VP$\rightarrow$VBD ; VBD$\rightarrow$occurred&  - \\
(4) Linearized derivation  & S\eor \ NP \ VP\eor \ DT \ NNS\eor \ No \ complications \ VBD\eor \ occurred&  73.8 \\
(5) POS/plain-text & DT  No  NNS  complications  VBD  occurred & 53.3  \\

\end{tabular}
\caption{Examples for proposed representations. Lengths are for the first 1M WAT English training sentences with BPE subwords \cite{sennrich2016subword}.}
          \label{tab:represent-lens}
 \end{table*}

\section{Ensembles of Syntax Models}
We wish to ensemble using models which generate linearized
constituency trees but these representations can be very long and difficult to model.  
We therefore propose a derivation-based representation which
is much more compact than a linearized parse tree (examples in Table~\ref{tab:represent-lens}).
Our linearized derivation representation ((4) in Table
\ref{tab:represent-lens}) consists of the derivation's right-hand side
tokens with an end-of-rule marker, \eor, marking the last non-terminal
in each rule. The original tree can be directly reproduced from the sequence,
so that structure information is maintained.  We map words to subwords
as described in Section~\ref{ss:data}.

\subsection{Delayed SGD Update Training for Long Sequences}
We suggest a training strategy for the Transformer model
\cite{vaswani2017attention} which gives improved performance for long
sequences, like syntax representations, without requiring additional
GPU memory. The Tensor2Tensor framework \cite{tensor2tensor} defines
batch size as the number of tokens per batch, so batches will
contain fewer sequences if their average length increases.  During NMT
training, by default, the gradients used to update model parameters
are calculated over individual batches.  A possible consequence is that 
batches containing fewer sequences per update  
may have `noisier' estimated gradients than batches with more sequences.

Previous research has used very large batches to improve training
convergence while requiring fewer model updates
\cite{smith2017dontdecay,neishi2017bag}. However, with such large
batches the model size may exceed available GPU memory. Training on
multiple GPUs is one way to increase the amount of data used
to estimate gradients, but it requires significant resources. Our
strategy avoids this problem by using delayed SGD updates. We
accumulate gradients over a fixed number of batches before using the
accumulated gradients to update the
model\footnote{\url{https://github.com/fstahlberg/tensor2tensor}}. This
lets us effectively use very large batch sizes without requiring
multiple GPUs.
\subsection{Ensembling Representations}

\begin{figure*}[h!]
\centering
\includegraphics[height=3.8cm]{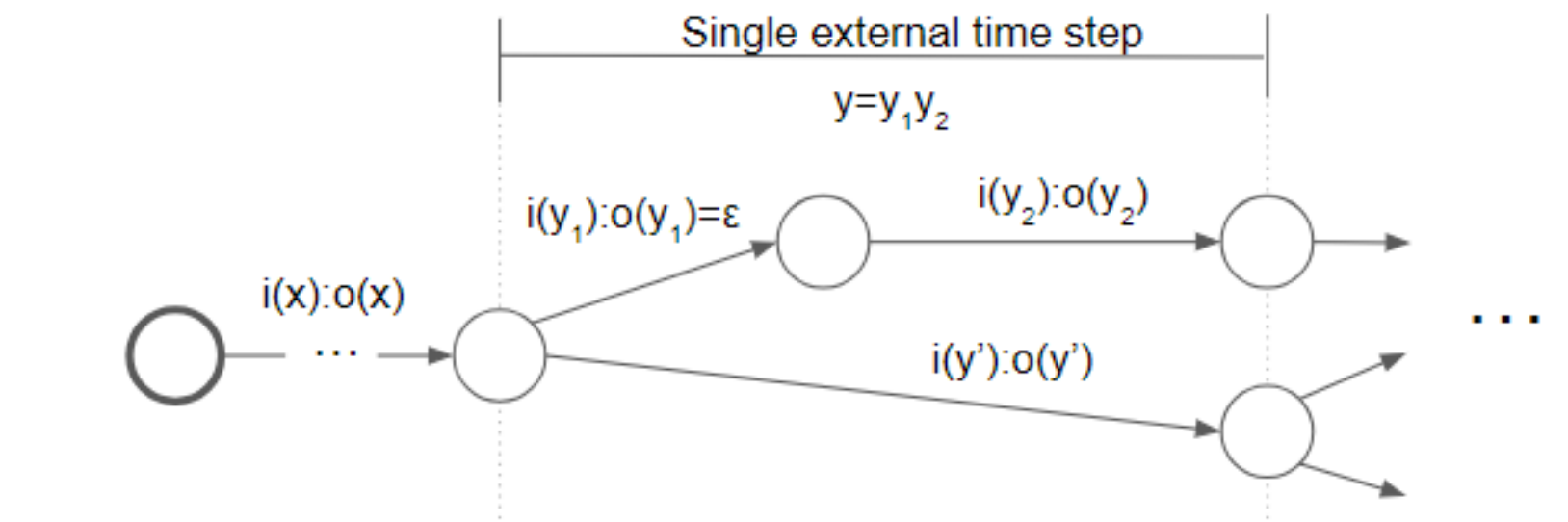}
\caption{Transducer mapping internal to external representations. A partial hypothesis might be $\mathsf{o(xy_2)}$ in the external representation and $\mathsf{i(xy_1y_2)}$ in the internal representation.}
\label{fig:fst-ensemble}
\end{figure*}

Table \ref{tab:represent-lens} shows several different representations
of the same hypothesis.  To formulate an ensembling decoder over pairs
of these representations, we assume we have a transducer $T$ that maps
from one representation to the other
representation.  The complexity of the transduction depends on the
representations.  Mapping from word to BPE representations is
straightforward, and mapping from (linearized) syntax to plain-text 
simply deletes non-terminals.
Let ${\mathcal P}$ be the paths in
$T$ leading from the start state to any final state.
A path $p \in {\mathcal P}$ maps an {\em internal representation} $i(p)$ to an {\em external representation} $o(p)$. 

The ensembling decoder produces  external representations.
Two NMT systems are trained, one for each representation, giving
models $P_{i}$ and $P_{o}$.   An ideal equal-weight
ensembling of $P_i$ and $P_o$ yields 
\begin{align}
p^\ast =\argmax_{p\in {\mathcal P}} \, P_{i}(i(p)) \,  P_{o}(o(p))\end{align}
with $o(p^\ast)$ as the external representation of the translation.

In practice, beam decoding is performed in the external
representation, i.e. over projections of paths in ${\mathcal P}$~\footnote{See
  the \texttt{tokenization} wrappers in
  \url{https://github.com/ucam-smt/sgnmt}}. 
Let $h = h_1 \ldots h_j$ be a partial hypothesis in the output representation.
The set of partial paths yielding $h$ are:
\begin{align}
&M(h) = \\
\nonumber &\{(x,y) | xyz \in {\mathcal P}, o(x)=h_{<j}, o(xy)=h\} \end{align}
Here $z$ is the path suffix. The ensembled score of $h$ is then:
\begin{align} P(h_j | h_{<j}) = &P_{o}(h_j | h_{<j}) \times\\
\nonumber &\max_{(x,y) \in M(h)} P_{i}(i(y)|i(x)) \end{align}
The $\max$ performed for each partial hypothesis $h$ is itself approximated by a beam search. This leads to an outer beam search over external representations with inner beam searches for the best matching internal representations. As search proceeds, each model score is updated separately with its appropriate representation. Symbols in the internal representation are consumed as needed to stay synchronized with the external representation, as illustrated in  Figure \ref{fig:fst-ensemble};  epsilons are consumed with a probability of 1.

\section{Experiments}
\label{ss:data}
\begin{table*}[h!]
\centering
\small
\begin{tabular}{p{1.6cm} | p{12cm}}
    \hline
   Reference & low - energy electron microscope ( LEEM ) and photoelectron microscope ( PEEM ) were attracted attention as new surface electron microscope .\\
 Plain BPE & low energy electron microscope ( LEEM ) and photoelectron microscope ( PEEM ) are noticed as new surface electron microscope . \\
Linearized derivation & low-energy electron microscopy ( LEEM ) and photoelectron microscopy ( PEEM ) are attracting attention as new surface electron microscopes .\\
    \hline
    \end{tabular}
    \caption{Sample generated translations from individual models}
        \label{tab:syntax-examples}
\end{table*}

We first explore the effect of our delayed SGD update training scheme on single models, contrasting updates every batch with accumulated updates every 8 batches. 
To compare target representations we train Transformer models with target representations (1), (2), (4) and (5) shown in Table \ref{tab:represent-lens}, using delayed SGD updates every 8 batches.  We decode with individual models and two-model ensembles, comparing results for single-representation and multi-representation ensembles. Each multi-representation ensemble consists of the plain BPE model and one other individual model. 

All Transformer architectures are Tensor2Tensor's base Transformer model \cite{tensor2tensor} with a batch size of 4096. In all cases we decode using SGNMT \citep{stahlberg2017sgnmt} with beam size 4, using the average of the final 20 checkpoints. For comparison with earlier target syntax work, we also train two RNN attention-based seq2seq models \citep{bahdanau15jointly} with normal SGD to produce plain BPE sequences and linearized derivations. For these models we use embedding size 400, a single BiLSTM layer of size 750, and batch size 80.

We report all experiments for Japanese-English, using the first 1M training sentences of the Japanese-English ASPEC data \cite{aspec}. All models use plain BPE Japanese source sentences. English constituency trees are obtained using CKYlark \cite{oda2015ckylark}, with words replaced by BPE subwords. We train separate Japanese (lowercased) and English (cased) BPE vocabularies on the plain-text, with 30K merges each. 
Non-terminals are included as separate tokens. The linearized derivation uses additional tokens for non-terminals with \eor. 

\subsection{Results and Discussion}
Our first results in Table~\ref{tab:transformer-multibatch} show that large batch training can significantly improve the performance of single Transformers, particularly when trained to produce longer sequences.  Accumulating the gradient over 8 batches of size 4096 gives a 3 BLEU improvement for the linear derivation model. It has been suggested that decaying the learning rate can have a similar effect to large batch training \citep{smith2017dontdecay}, but reducing the initial learning rate by a factor of 8 alone did not give the same improvements.

\begin{table}[!h]
\centering
\small
\begin{tabular}{ p{2cm}| p{1.2cm} | p{1cm} |p{0.8cm} }
    \hline
    Representation & Batches / update & Learning rate & Test BLEU  \\
    \hline
    \multirow{ 3}{*}{Plain BPE} & 1 & 0.025 &27.5\\
       & 1 & 0.2 &27.2\\
       &8 & 0.2 & 28.9\\
     \hline
    \multirow{ 3}{*}{\parbox[t]{2cm}{Linearized derivation}} &  1&  0.025 & 25.6\\
    &  1& 0.2 & 25.6\\
     & 8 & 0.2 & 28.7\\
    \hline
    \end{tabular}
    \caption{Single Transformers trained to convergence on 1M WAT Ja-En, batch size 4096}
        \label{tab:transformer-multibatch}
\end{table}

\begin{table}[!h]
\centering
\small
\begin{tabular}{p{1.5cm} | >{\raggedright\arraybackslash}p{2.7cm} | p{0.7cm}|p{0.6cm}}
    \hline
   	Architecture & Representation &  Dev BLEU & Test BLEU\\
    \hline
    Seq2seq (8-model ensemble) & Best WAT17 result
    \cite{morishita2017ntt} & - & 28.4 \\
    \hline
     \multirow{ 2}{*}{Seq2seq} & Plain BPE & 21.6 & 21.2 \\
    & Linearized derivation & 21.9 & 21.2 \\ 
    \hline
   \multirow{4}{*}{Transformer} & Plain BPE & 28.0 & 28.9 \\
   & Linearized tree & 28.2 & 28.4 \\
   & Linearized derivation & 28.5 & 28.7 \\
   & POS/BPE & 28.5 & 29.1 \\
    \end{tabular}
    \caption{Single models on Ja-En. Previous evaluation result included for comparison.}
        \label{tab:results-single}
\end{table}

\begin{table}[!h]
\centering
\small
\begin{tabular}{>{\raggedright\arraybackslash}p{2.7cm} >{\raggedright\arraybackslash}p{2.7cm} | p{0.6cm} }
    \hline
   External representation & Internal representation &  Test BLEU\\
    \hline
         Plain BPE  & Plain BPE &29.2 \\
     Linearized derivation & Linearized derivation  &28.8 \\
\hline
     Linearized tree & Plain BPE &28.9 \\
    Plain BPE &Linearized derivation &  28.8 \\
    Linearized derivation &Plain BPE &  29.4$^\dagger$ \\
     POS/BPE & Plain BPE & 29.3$^\dagger$ \\
     Plain BPE & POS/BPE &  29.4$^\dagger$\\
    \end{tabular}
    \caption{Ja-En Transformer ensembles:  $\dagger$ marks significant improvement on plain BPE baseline shown in Table \ref{tab:results-single} ($p<0.05$ using bootstrap resampling \citep{koehn2007moses}).}
        \label{tab:results-ensemble}
\end{table}

Our plain BPE baseline (Table \ref{tab:results-single}) outperforms the current best system on WAT Ja-En, an 8-model ensemble \cite{morishita2017ntt}. Our syntax models achieve similar results despite producing much longer sequences. Table \ref{tab:transformer-multibatch} indicates that large batch training is instrumental in this. 

We find that RNN-based syntax models can equal plain BPE models as in \citet{aharoni2017towards}; \citet{eriguchi2017learning} use syntax for a 1 BLEU improvement on this dataset, but over a much lower baseline. Our plain BPE Transformer outperforms all syntax models except POS/BPE. More compact syntax representations perform better, with POS/BPE outperforming linearized derivations, which outperform linearized trees. 

Ensembles of two identical models trained with different seeds only slightly improve over the single model (Table \ref{tab:results-ensemble}). However, an ensemble of models producing plain BPE and linearized derivations improves by 0.5 BLEU over the plain BPE baseline. 

By ensembling syntax and plain-text we hope to benefit from their complementary strengths. To highlight these, we examine hypotheses generated by the plain BPE and linearized derivation models. We find that the syntax model is often more grammatical, even when the plain BPE model may share more vocabulary with the reference (Table \ref{tab:syntax-examples}).

In ensembling plain-text with a syntax external representation we observed that in a small proportion of cases non-terminals were over-generated, due to the mismatch in target sequence lengths.  Our solution was to penalise scores of non-terminals under the syntax model by a constant factor.

It is also possible to constrain decoding of linearized trees and
derivations to well-formed outputs. However, we found that this gives
little improvement in BLEU over unconstrained decoding although it
remains an interesting line of research.
\section{Conclusions}
We report strong performance with individual models that meets or improves over the
recent best WAT Ja-En ensemble results.  We train these models using a
delayed SGD update training procedure that is especially effective for
the long representations that arise from including target language
syntactic information in the output.  We further improve on the
individual results via a decoding strategy allowing ensembling of
models producing different output representations, such as subword
units and syntax.  We propose these techniques as practical approaches
to including target syntax in NMT.
\section*{Acknowledgments}
This work was supported by EPSRC grant EP/L027623/1.

\bibliographystyle{acl_natbib}
\bibliography{refs}

\begin{thebibliography}{19}
\expandafter\ifx\csname natexlab\endcsname\relax\def\natexlab#1{#1}\fi

\bibitem[{Aharoni and Goldberg(2017)}]{aharoni2017towards}
Roee Aharoni and Yoav Goldberg. 2017.
\newblock Towards string-to-tree neural machine translation.
\newblock In \emph{Proceedings of the 55th Annual Meeting of the Association
  for Computational Linguistics (Volume 2: Short Papers)}, volume~2, pages
  132--140.

\bibitem[{Bahdanau et~al.(2015)Bahdanau, Cho, and Bengio}]{bahdanau15jointly}
Dzmitry Bahdanau, Kyunghyun Cho, and Yoshua Bengio. 2015.
\newblock Neural machine translation by jointly learning to align and
  translate.
\newblock In \emph{ICLR}.

\bibitem[{Dyer et~al.(2016)Dyer, Kuncoro, Ballesteros, and
  Smith}]{dyer2016recurrent}
Chris Dyer, Adhiguna Kuncoro, Miguel Ballesteros, and Noah~A Smith. 2016.
\newblock Recurrent neural network grammars.
\newblock In \emph{Proceedings of NAACL-HLT}, pages 199--209.

\bibitem[{Eriguchi et~al.(2017)Eriguchi, Tsuruoka, and
  Cho}]{eriguchi2017learning}
Akiko Eriguchi, Yoshimasa Tsuruoka, and Kyunghyun Cho. 2017.
\newblock Learning to parse and translate improves neural machine translation.
\newblock In \emph{Proceedings of the 55th Annual Meeting of the Association
  for Computational Linguistics (Volume 2: Short Papers)}, volume~2, pages
  72--78.

\bibitem[{Garmash and Monz(2016)}]{garmash2016ensemble}
Ekaterina Garmash and Christof Monz. 2016.
\newblock Ensemble learning for multi-source neural machine translation.
\newblock In \emph{Proceedings of COLING 2016, the 26th International
  Conference on Computational Linguistics: Technical Papers}, pages 1409--1418.

\bibitem[{Hokamp(2017)}]{hokamp2017ensembling}
Chris Hokamp. 2017.
\newblock Ensembling factored neural machine translation models for automatic
  post-editing and quality estimation.
\newblock In \emph{Proceedings of the Second Conference on Machine
  Translation}, pages 647--654.

\bibitem[{Koehn et~al.(2007)Koehn, Hoang, Birch, Callison-Burch, Federico,
  Bertoldi, Cowan, Shen, Moran, Zens et~al.}]{koehn2007moses}
Philipp Koehn, Hieu Hoang, Alexandra Birch, Chris Callison-Burch, Marcello
  Federico, Nicola Bertoldi, Brooke Cowan, Wade Shen, Christine Moran, Richard
  Zens, et~al. 2007.
\newblock Moses: Open source toolkit for statistical machine translation.
\newblock In \emph{Proceedings of the 45th annual meeting of the ACL on
  interactive poster and demonstration sessions}, pages 177--180. Association
  for Computational Linguistics.

\bibitem[{Morishita et~al.(2017)Morishita, Suzuki, and
  Nagata}]{morishita2017ntt}
Makoto Morishita, Jun Suzuki, and Masaaki Nagata. 2017.
\newblock {NTT} neural machine translation systems at {WAT} 2017.
\newblock In \emph{Proceedings of the 4th Workshop on Asian Translation
  (WAT2017)}, pages 89--94.

\bibitem[{Nadejde et~al.(2017)Nadejde, Reddy, Sennrich, Dwojak,
  Junczys-Dowmunt, Koehn, and Birch}]{nadejde2017predicting}
Maria Nadejde, Siva Reddy, Rico Sennrich, Tomasz Dwojak, Marcin
  Junczys-Dowmunt, Philipp Koehn, and Alexandra Birch. 2017.
\newblock Predicting target language {CCG} supertags improves neural machine
  translation.
\newblock In \emph{Proceedings of the Second Conference on Machine
  Translation}, pages 68--79.

\bibitem[{Nakazawa et~al.(2016)Nakazawa, Yaguchi, Uchimoto, Utiyama, Sumita,
  Kurohashi, and Isahara}]{aspec}
Toshiaki Nakazawa, Manabu Yaguchi, Kiyotaka Uchimoto, Masao Utiyama, Eiichiro
  Sumita, Sadao Kurohashi, and Hitoshi Isahara. 2016.
\newblock {ASPEC}: Asian scientific paper excerpt corpus.
\newblock In \emph{Proceedings of the Ninth International Conference on
  Language Resources and Evaluation (LREC)}, pages 2204--2208. European
  Language Resources Association (ELRA).

\bibitem[{Neishi et~al.(2017)Neishi, Sakuma, Tohda, Ishiwatari, Yoshinaga, and
  Toyoda}]{neishi2017bag}
Masato Neishi, Jin Sakuma, Satoshi Tohda, Shonosuke Ishiwatari, Naoki
  Yoshinaga, and Masashi Toyoda. 2017.
\newblock A bag of useful tricks for practical neural machine translation:
  Embedding layer initialization and large batch size.
\newblock In \emph{Proceedings of the 4th Workshop on Asian Translation
  (WAT2017)}, pages 99--109.

\bibitem[{Nivre(2004)}]{nivre2004incrementality}
Joakim Nivre. 2004.
\newblock Incrementality in deterministic dependency parsing.
\newblock In \emph{Proceedings of the Workshop on Incremental Parsing: Bringing
  Engineering and Cognition Together}, pages 50--57. Association for
  Computational Linguistics.

\bibitem[{Oda et~al.(2015)Oda, Neubig, Sakti, Toda, and
  Nakamura}]{oda2015ckylark}
Yusuke Oda, Graham Neubig, Sakriani Sakti, Tomoki Toda, and Satoshi Nakamura.
  2015.
\newblock Ckylark: A more robust {PCFG-LA} parser.
\newblock In \emph{Proceedings of the 2015 Conference of the North American
  Chapter of the Association for Computational Linguistics: Demonstrations},
  pages 41--45.

\bibitem[{Sennrich et~al.(2016)Sennrich, Haddow, and
  Birch}]{sennrich2016subword}
Rico Sennrich, Barry Haddow, and Alexandra Birch. 2016.
\newblock Neural machine translation of rare words with subword units.
\newblock In \emph{Proceedings of the 54th Annual Meeting of the Association
  for Computational Linguistics (Volume 1: Long Papers)}, volume~1, pages
  1715--1725.

\bibitem[{Smith et~al.(2017)Smith, Kindermans, and Le}]{smith2017dontdecay}
Samuel~L Smith, Pieter-Jan Kindermans, and Quoc~V Le. 2017.
\newblock Don't decay the learning rate, increase the batch size.
\newblock \emph{arXiv preprint arXiv:1711.00489}.

\bibitem[{Stahlberg et~al.(2017)Stahlberg, Hasler, Saunders, and
  Byrne}]{stahlberg2017sgnmt}
Felix Stahlberg, Eva Hasler, Danielle Saunders, and Bill Byrne. 2017.
\newblock {SGNMT}--a flexible {NMT} decoding platform for quick prototyping of
  new models and search strategies.
\newblock In \emph{Proceedings of the 2017 Conference on Empirical Methods in
  Natural Language Processing: System Demonstrations}, pages 25--30.

\bibitem[{Vaswani et~al.(2018)Vaswani, Bengio, Brevdo, Chollet, Gomez, Gouws,
  Jones, Kaiser, Kalchbrenner, Parmar, Sepassi, Shazeer, and
  Uszkoreit}]{tensor2tensor}
Ashish Vaswani, Samy Bengio, Eugene Brevdo, Francois Chollet, Aidan~N. Gomez,
  Stephan Gouws, Llion Jones, \L{}ukasz Kaiser, Nal Kalchbrenner, Niki Parmar,
  Ryan Sepassi, Noam Shazeer, and Jakob Uszkoreit. 2018.
\newblock Tensor2tensor for neural machine translation.
\newblock \emph{CoRR}, abs/1803.07416.

\bibitem[{Vaswani et~al.(2017)Vaswani, Shazeer, Parmar, Uszkoreit, Jones,
  Gomez, Kaiser, and Polosukhin}]{vaswani2017attention}
Ashish Vaswani, Noam Shazeer, Niki Parmar, Jakob Uszkoreit, Llion Jones,
  Aidan~N Gomez, {\L}ukasz Kaiser, and Illia Polosukhin. 2017.
\newblock Attention is all you need.
\newblock In \emph{Advances in Neural Information Processing Systems}, pages
  6000--6010.

\bibitem[{Wu et~al.(2017)Wu, Zhang, Yang, Li, and Zhou}]{wu2017sequence}
Shuangzhi Wu, Dongdong Zhang, Nan Yang, Mu~Li, and Ming Zhou. 2017.
\newblock Sequence-to-dependency neural machine translation.
\newblock In \emph{Proceedings of the 55th Annual Meeting of the Association
  for Computational Linguistics (Volume 1: Long Papers)}, volume~1, pages
  698--707.

\end{thebibliography}

\end{document}